# AN FPGA-BASED PARALLEL ARCHITECTURE FOR FACE DETECTION USING MIXED COLOR MODELS

LUO Tao

*School of Computer Science and Technology，Tianjin University,*
*Tianjin Key Laboratory of Cognitive Computing and Application,*
*Tianjin 300072，China*
*luo_tao@tju.edu.cn*

Zaifeng Shi

*School of Electronic Information Engineering，Tianjin University,*
*Tianjin 300072，China*
*shizaifeng@tju.edu.cn*

In this paper, a reliable method for detecting human faces in color images is proposed. This system firstly detects skin color in YCgCr and YIQ color space, then filters binary texture and the result is morphological processed, finally converts skin tone to the preferred skin color configured by users in YIQ color space. The real-time adjusting circuit is implemented and some of simulation results are given out. Experimental results demonstrate that the method has achieved high rates and low false positives , another advantage is its simplicity and minor computational costs.

## 1. Introduction

Human face perception is currently an active research area in the computer vision community. Human face localization and detection is often the first step in applications such as video surveillance, human computer interface, face recognition and image database management.

For detecting face there are various algorithms including skin color based algorithms. Color is an important feature of human faces. Using skin-color as a feature for tracking a face has several advantages.

Skin color has proven to be a useful and robust cue for face detection, localization and tracking in [1,2].Human skin color has been considered as a distinguishing and effective feature in face detection, taking into account that the major differences between people having different skin color lie mostly in their intensity rather than their chrominance [3,4]. For the classification of the pixels in an image into skin and non-skin regions, several color spaces have been proposed, such as RGB, XYZ, CIE-Lab, HSV or YCbCr[5-8]. Zhang[9] presents a detection method based on uniting YCgCb color space with YCgCr color space, Color spaces which work with separated luminance and chrominance components, like HSV or YCbCr, seem to be more appropriate for face detection.

In this paper, a reliable method for detecting human faces in color images is introduced. The system detects skin color in YCgCr and YIQ color space, then filters binary texture and the result is morphological processed, finally converts skin tone to the preferred skin color configured by users in YIQ color space. The paper propose a parallel hardware architecture. The proposed architecture is designed using VHDL and implemented on a Virtex-5 FPGA for prototyping and evaluation. Experimental results demonstrate that our system has achieved high rates and low false positives and another advantage of this method is its simplicity and minor computational costs.

## 2. Proposed Methodology

The main steps employed for the face detection process in a color image is shown in Fig. 1.

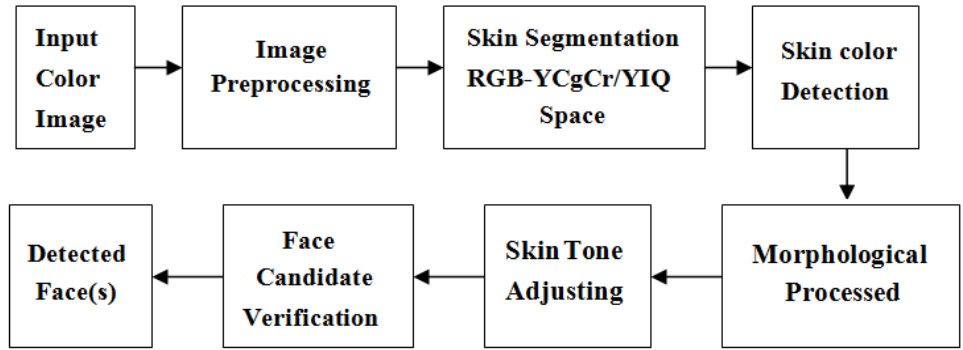

Fig 1. The procedure of proposed methodology

### 2.1. Color-Space Selection

YCbCr is most commonly used model in digital video because it has the smallest overlap between the skin and non skin data in under various illumination conditions. It belongs to the family of television transmission color space. These color spaces separate RGB into luminance and chrominance information and are very useful in compression applications however the specification of colors is somewhat unintuitive. In the YCbCr color space, Y is the luminance component and Cb and Cr are the blue-difference and red-difference chroma components.

YCgCr color space was derived from YCgCr color space. The YCgCr color space is a variation of YCbCr color space that uses the color difference (G-Y) instead of (B-Y). YCgCr components can be derived from the following matrix expression:

$$\begin{bmatrix} Y \\ Cg \\ Cr \end{bmatrix} = \begin{bmatrix} 16 \\ 128 \\ 128 \end{bmatrix} + \begin{bmatrix} 65.481 & 128.553 & 24.966 \\ -81.085 & 112 & -30.915 \\ 112 & -93.786 & -18.214 \end{bmatrix} \begin{bmatrix} R \\ G \\ B \end{bmatrix} \quad (1)$$

YIQ color model belongs to the family of television transmission color models. The main advantages of this format is that grayscale information is separated from color data, so the same signal can be used for both color and black and white sets. In this color model, image data consists of three components: luminance (Y) which represents grayscale information, while hue (I) and saturation (Q) represents the color information.

$$\begin{bmatrix} Y \\ I \\ Q \end{bmatrix} = \begin{bmatrix} 0.299 & 0.587 & 0.114 \\ 0.596 & -0.274 & -0.322 \\ 0.212 & -0.523 & -0.311 \end{bmatrix} \begin{bmatrix} R \\ G \\ B \end{bmatrix} \quad (2)$$

In this paper, ig skin color model. Because the chrominance components such as Cg, Cr are separated from the luminance Y. So an effective use of the chrominance information for modeling human skin color can be achieved in these color space. Secondly, this format is typically used in video coding, and therefore the use of the same format for segmentation will avoid the extra computation required.

**2.2. Skin-Color Model**

Zhang[9] presents experiments of 1010 skin tone pixels which are in different ages and body areas. Finally, he builds a parallelogram model in Cg-Cr color space for skin tone detection. The model for the skin tone in the transformed Cg-Cr space is described in expression (3), and this model is used in this paper.

$$\begin{cases} Cr \in [-Cg+260 \quad -Cg+280] \\ Cg \in [85 \quad 135] \end{cases} \quad (3)$$

In this paper, skin color detected by uniting YCgCr color space with YIQ color space. It is found that a rectangle model could approximately represent the skin tone distribution. The rectangle model for the skin tone in the transformed I-Q space is described in expression (4), it is more simple for hardware implementation.

$$\begin{cases} I \in [15 \quad 90] \\ Q \in [-20 \quad 10] \end{cases} \quad (4)$$

**2.3 Morphological opening**

For the sake of removing the noise and the background pixels, the binary skin detected result is morphological processed. Morphological opening removes completely regions of an object that cannot contain the structuring element, smoothes object contours, breaks thin connections, and removes thin protrusions. The goal is to end up with a mask image that can be applied to the input image to yield skin color regions without noise and clutter[10].

Morphological opening is performed to remove very small objects from the image while preserving the shape and size of larger objects in the image. The definition of a morphological opening of an image is an erosion followed by a dilation, using the same structuring element for both operations. The morphological opening of A by B, firstly erosion of A by B, followed by dilation of the result by B.

In this paper, the structuring element B is chosen as circle with radius of 2. Fig2 illustrates how morphological opening works, Fig2(b) shows an original binary image which shows only the 1s, the gray region indicates an element B has been found in the image and the centre pixel will be a 1 in result of the erosion; Fig2(c) shows the erosion result of the original image, the gray region indicates dilation following erosion with a 1 at the centre; Fig2(d) shows the dilation result of the eroded image, which is also the result of morphological opening of original image by element B. As the result shown, morphological opening breaks thin connections and smoothes object contours.

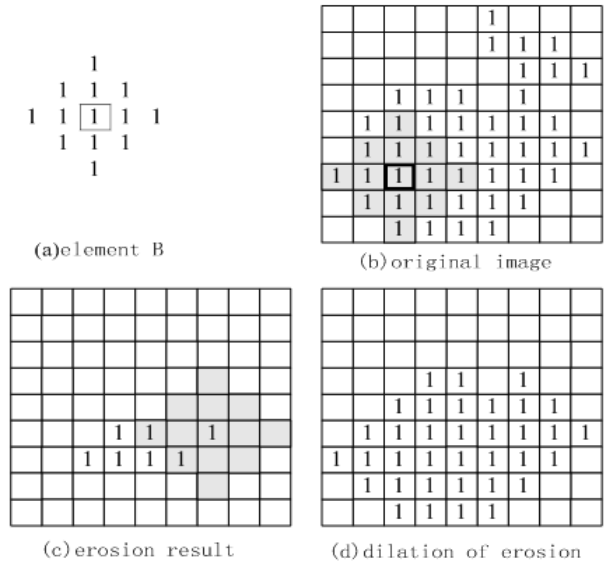

Fig2. Illustration of morphological opening

The elementary result of skin region detection is logic AND between the binary images by the parallelogram model in YCgCb color space and by the rectangle model in YIQ color space. Then the binary detected result subtracts the texture and is morphological processed.

**2.4 User-configurable skin tone adjusting**

The pixels are converted to YCgCb and YIQ color space while skin tone region is detected, so it will be efficient to adjust skin tone in YIQ color space. Since different people have different views on beautiful skin color, a user-configurable skin tone adjusting is proposed. It allows user to select directions and range of the skin tone variation. Fig3 shows The diagram of skin tone adjusting, there roughly are four directions for skin tone variation:

**steps:** 1. both I and Q variable are increased;
    2. I variable is increased while Q is decreased;
    3. both I and Q variable are decreased;
    4. I variable is decreased while Q is increased.

These four variations respectively indicate the skin tone is moved toward red, yellow, green and magenta, Fig4 shows the results of four variations. The ranges of the variations are percents of original I and Q values, which are configured by user. All of what can be configured is two 16-bit signed integer registers, which indicate the range of the variation from -100% to 100%. For example, if user select a range of -18%, the register should be configured as "16'b1110 1000 1111 0110" or "16'd-5898". I and Q of the pixels in detected skin tone region are adjusted as expression (5), in which I_range and Q_range indicate the signed range of the variation.

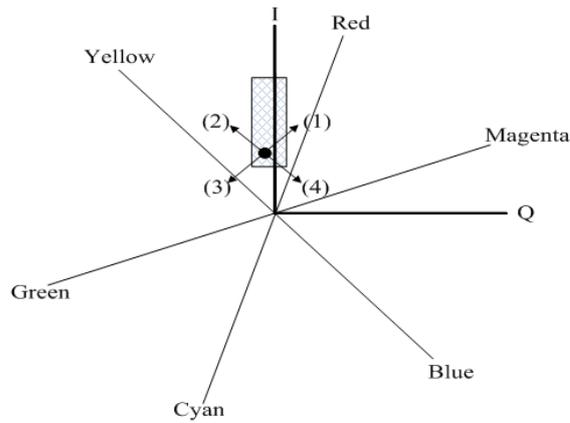

Fig3. The diagram of skin tone adjusting

$$\begin{cases} I\_out = I + I \times I\_range \\ Q\_out = Q + Q \times Q\_range \end{cases} \quad (5)$$

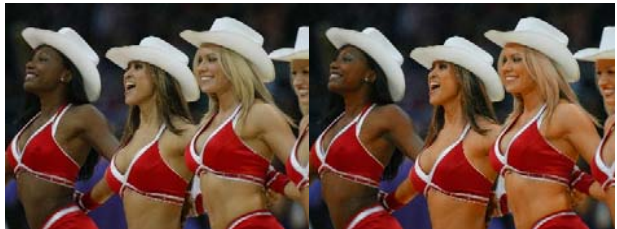

(a) original image　　（b）step(1)

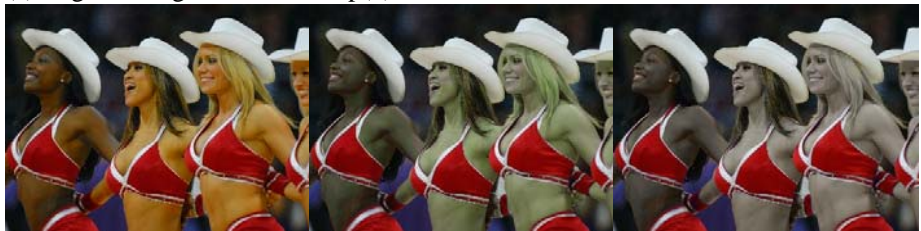

step(2)　　　step(3)　　　step(4)
Fig4. The results of four directions for skin tone variation

**3. Configurable real-time adjusting circuit and the results**

The hardware architecture is shown in Fig5. A pipeline circuit is used to process each input pixel in time. A 3-line parallel video_in data are input every line cycle because of 3×3 structuring element being used in morphological gradient.

1.The Gamut conversion module converts RGB data to YIQ and YCgCr formats to ensure that results of I-Q and Cg-Cr skin detection are synchronized with the result of texture detection. The skin detection module uses IQ data and CgCr data to decide whether the pixel accords with the rectangle model described in expression (3) and (4), on the other hand YIQ data are simultaneously sent to YIQ data buffer to be synchronized with the result of the morphological opening. The texture detection module on the one hand transforms 3-line parallel input data to gray data and thresholds the value of the morphological gradient to gain the binary texture.on the other hand, the module judges that the pixel whether satisfying the inequality of R<80, G<80, B<80 and whether not satisfying the inequality of R>G>B when R<230, G<230 and B<230. The output of the texture detection module is logic OR between the two results. A 3-input AND logic is used after the detections, the three input are respectively result of I-Q skin detection, result of Cg-Cr skin detection and the reverse of the texture detection result.

2.The morphological opening module receives the result of the AND logic and buffers it for following erosion and dilation.

3.The skin tone adjusting module receives the result of morphological opening, gets YIQ data from the buffer and obtains the directions and range of the skin tone variation from the user register module, then adjusts the pixels judged as skin pixels and keeps the original YIQ data if the pixel is judged as a non-skin pixel.

4.Finally, the pixel data are transformed from the YIQ color space back to RGB color space and are output.

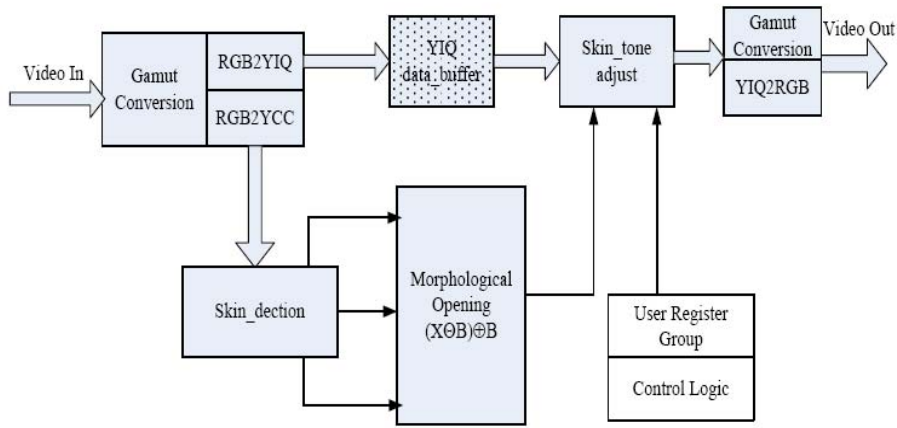

Fig5. Hardware architecture of skin tone adjusting

Fig6 shows the performance of the skin tone adjusting using the recommended variation range. The left of each group is original image, the right one is skin tone adjusted image. As the image shown, the skins of the results look rosy and glow, the figure in the adjusted image seem healthier and more energetic.

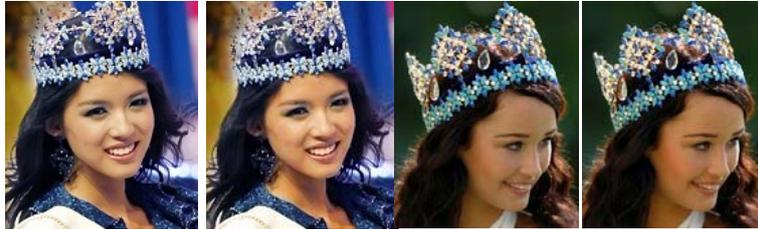

Fig6. Performance of the skin tone adjusting using the recommended variation range

The entire circuit has been implemented using VHDL. We use a camera with a 33 MHz clock frequency for the implementation. The hardware module is synthesized by the Synplify logic synthesis tool from Synplicity. We evaluated the power dissipated by our FPGA implementation in the Xilinx Power Estimator. The total power dissipated was estimated at 8.9W. Table 1 shows the device utilization summary for the implementation of proposed face detection system on an Virtex-5 FPGA device by the Xilinx ISE. Experimental results demonstrate that the circuit has the potential of achieving speeds far higher than real-time video processing requirements, even at low clock speeds.

Table.1  Device utilization

| Resources | Used | Available | Utility Ratio |
|---|---|---|---|
| **Slices** | **11406** | **34560** | **33%** |
| **CLB Flip-Flops** | **82985** | **138240** | **60%** |
| **Block RAM (kbits)** | **536** | **6912** | **8%** |
| Averaged Running Time 8.137sec ||||

4. **Experimental Results and Analyses**

In the experiment, we choose 800 normal image for detection. Table 2 shows the comparison of four algorithms. The experimental results show that the proposed method can exclude some background pixels which are similar to skin tone. In these binary images, we use white to express the skin color and black for the non-skin color. As shown in Fig7, The experiments show the satisfactory results have been achieved by using proposed method Moreover, skin of all race can be detected precisely, this ensures that further adjusting is meaningful.

Table 2. Comparison Chart of the Algorithms

| Criterion | RGB Color Space | YcbCr Color Space | HSV Color Space | Proposed |
|---|---|---|---|---|
| Normal images | 69.31% | 77.90% | 84.51% | 86.12% |

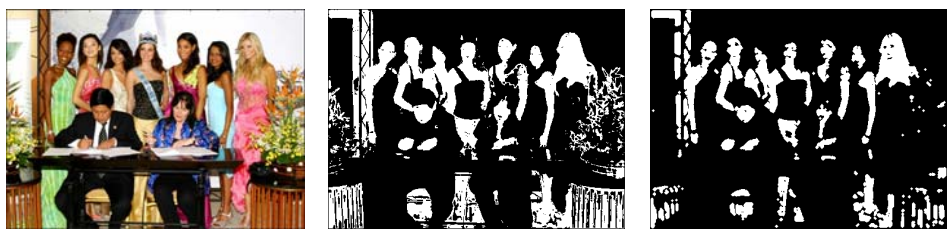

(a) original image     (b) elementary result     (c) final result

Fig7. The results of proposed method

## 5. Conclusion

A novel effective image representations for face recognition. is proposed. In order to raise the accuracy of skin color region detection, the method units YCgCr color space with YIQ color to detect the skin tone region, binary texture and two conditions are used to filter the non-skin pixels, the binary result is finally morphological opening processed. Instead of searching for fixed preferred skin color, a user-configurable skin tone adjusting method is proposed. The system converts skin tone to the preferred skin color configured by users in YIQ color space and outputs RGB data converted back. The real-time adjusting circuit is implemented and some of simulation results are given out.

## 6. Acknowledgments

This work was supported by National Natural Science Foundation of China grants 61306070.